\documentclass[twoside,11pt]{article}
\usepackage[preprint]{jmlr2e}

\usepackage[normalem]{ulem}
\usepackage{natbib}

\usepackage[colorinlistoftodos]{todonotes}
\usepackage{graphicx}

\newcommand{\ignore}[1]{}

\firstpageno{1}

\begin{document}
\thispagestyle{empty}
\title{From Machine Learning to Robotics: Challenges and Opportunities for Embodied Intelligence}

\author{\name Nicholas Roy \email nickroy@csail.mit.edu \\
\addr Massachusetts Institute of Technology, Cambridge, MA USA 
\AND 
\name Ingmar Posner \email ingmar@robots.ox.ac.uk \\
\addr University of Oxford, Oxford, UK 
\AND 
Tim Barfoot \email tim.barfoot@utoronto.ca \\
\addr University of Toronto, Toronto, ON Canada
\AND 
Philippe Beaudoin \email phil@mywaverly.com \\
\addr Waverly, Montreal, QC Canada 
\AND 
Yoshua Bengio \email yoshua.bengio@mila.quebec \\
\addr Mila, University of Montreal, Montreal, QC Canada
\AND 
Jeannette Bohg \email bohg@stanford.edu \\
\addr Stanford University, Stanford, CA USA 
\AND 
\name Oliver Brock \email oliver.brock@tu-berlin.de\\
\addr Technische Universit\"{a}t Berlin, Berlin, Germany 
\AND 
\name Isabelle D\'epatie \email isabelle.depatie@elementai.com\\
\addr Element AI -- A ServiceNow Company, Montreal, QC Canada
\AND 
\name Dieter Fox \email fox@cs.washington.edu \\
\addr University of Washington, Seattle, WA USA 
\AND 
\name Dan Koditschek \email kod@seas.upenn.edu \\
\addr University of Pennsylvania, Philadelphia, PA USA 
\AND 
\name Tom\'as Lozano-P\'erez  \email tlp@csail.mit.edu \\
\addr Massachusetts Institute of Technology, 
Cambridge, MA USA 
\AND 
\name Vikash Mansinghka \email vkm@mit.edu \\
\addr Massachusetts Institute of Technology, 
Cambridge, MA USA 
\AND 
\name Christopher Pal \email christopher.pal@servicenow.com \\
\addr Element AI -- A ServiceNow Company, Montreal, QC Canada
\AND 
\name Blake Richards \email blake.richards@mila.quebec \\
\addr Mila, McGill University, 
Montreal, QC Canada 
\AND 
\name Dorsa Sadigh \email dorsa@cs.stanford.edu \\
\addr Stanford University, Stanford, CA USA 
\AND 
\name Stefan Schaal \email stefan.k.schaal@gmail.com\\
\addr [Google] X, 
Mountain View, CA USA 
\AND 
\name Gaurav Sukhatme \email gaurav@usc.edu \\
\addr University of Southern California, 
Los Angeles, CA USA 
\AND 
\name Denis Th\'erien \email denis@servicenow.com\\
\addr ServiceNow, Montreal, QC Canada
\AND 
\name Marc Toussaint \email toussaint@tu-berlin.de\\
\addr Technische Universit\"{a}t Berlin, 
Berlin, Germany 
\AND 
\name Michiel Van de Panne \email van@cs.ubc.ca\\
\addr University of British Columbia, 
Vancouver, BC Canada\\
\vspace{-.1in}
}

\editor{}

\maketitle
\vspace{-1in}

\clearpage 

\begin{abstract}
Machine learning has long since become a keystone technology, accelerating science and applications in a broad range of domains. Consequently, the notion of applying learning methods to a particular problem set has become an established and valuable modus operandi to advance a particular field. In this article we argue that such an approach does not straightforwardly extended to robotics --- or to \emph{embodied intelligence} more generally: systems which engage in a purposeful exchange of energy and information with a physical environment. In particular, the purview of embodied intelligent agents extends significantly beyond the typical considerations of main-stream machine learning approaches, which typically (i) do not consider operation under conditions significantly different from those encountered during training; (ii) do not consider the often substantial, long-lasting and potentially safety-critical nature of interactions during learning and deployment; (iii) do not require ready adaptation to novel tasks while at the same time (iv) effectively and efficiently curating and extending their models of the world through targeted and deliberate actions. In reality, therefore, these limitations result in learning-based systems which suffer from many of the same operational shortcomings as more traditional, engineering-based approaches when deployed on a robot outside a well defined, and often narrow operating envelope. Contrary to viewing embodied intelligence as another application domain for machine learning, here we argue that it is in fact a key driver for the advancement of machine learning technology. In this article our goal is to highlight challenges and opportunities that are specific to embodied intelligence and to propose research directions which may significantly advance the state-of-the-art in \emph{robot} learning.  


\end{abstract}

\section{Introduction}
\label{sec:introduction}
Robots and autonomous vehicles are being deployed with increasing frequency in an ever-increasing number of applications, from fleets of self-driving cars, to increasingly unpopulated factory environments to drones making commercial deliveries to people's houses. The expectation of the general public is that, more and more, robots are able to operate robustly and effectively, carrying out a range of tasks in a variety of complex domains. However, the reality is that today's robots are far from as robust, efficient and intelligent as they may seem. Today's robots are still brittle in the face of the unexpected, lack versatility, and are able to perform only a specific set of tasks within a very narrow set of operating conditions. It is an open secret that the vast majority of today's autonomous robots rely heavily on human supervision and intervention when deployed out in the world.

In the last ten years, there has been rapid progress on certain kinds of tasks in computer vision and natural language processing, driven by machine learning. For specific problems such as face recognition and machine translation, especially in the context of the web, engineered models have been no match for learned systems.  Robotics and embodied systems have derived some benefit from machine learning in machine vision, and there are specific capabilities which enable systems such as self-driving cars that would not exist without learning. However, machine learning appears to have run into many of the same problems of brittleness and lack of versatility as more traditional engineering approaches when deployed on a robot operating outside well-defined conditions such as the warehouse, factory floor, or carefully mapped streets in a city with an unvarying, sunny climate.

The challenge of developing embodied, physical robots that can learn reliably suggests that learning for robots and embodied agents may somehow be different from the domains where there has been greater operational success in using learning. In this article our goal is to highlight the challenges and opportunities that are specific to embodied intelligence and to propose research directions that may significantly advance the current state-of-the-art in robot learning.

\subsection*{The Challenges of Embodied Intelligence}

Although intelligent agents that inhabit a real physical body are often referred to as ``embodied'', there is not a consensus as to what precisely makes an intelligence ``embodied'' \citep{pfeifer2006body,cangelosi2015embodied,Howard2019,habitat19arxiv}. Whether embodiment requires a physical body or whether the term can also apply to agents acting purely in simulation is a matter of ongoing debate. We therefore define embodied intelligence as \emph{the purposeful exchange of energy and information with a physical environment} \citep{Koditschek2021}. We use the term ``physical environment'', rather than ``real world'' environment, in that a physical environment, simulated or otherwise, is one that enforces physical constraints of different kinds. In particular, our motivation lies in the field of robotics, where we are concerned with embodied agents which are equipped to act and interact in the real world and within the physical constraints imposed thereby.  It has long been conjectured that physical constraints present an embodied intelligent agent with a very different learning landscape than traditional AI agents, and the technical requirements for enabling these agents to learn are fundamentally different from the requirements of learning agents that do not interact with a physical world. 



Firstly, the fact that embodied agents exchange energy with the world implies that the agent's actions can have substantial and long-lasting effects, creating challenges of learning safely.  Often unbeknownst to the agent, parts of the world may carry an extremely high penalty of exploration.  Embodied agents must be aware of potentially catastrophic events that can fundamentally end any further learning or agency.  Embodied agents must also be aware of their own limits in terms of available energy and specific power, and be able to learn in real-time. Nature provides existence proofs of embodied agents that can learn to execute complex tasks from a tiny number of examples without endangering themselves or the environment, but artificial embodied intelligence has struggled to replicate this ability.  Behavioral studies of animal sensorimotor intelligence \citep{gallistel1990organization} and tool use \citep{seed2010animal} reveal capacities for spatial intelligence and problem solving that outstrip our best autonomous systems not only in robustness and accuracy but also data efficiency and power efficiency.


Secondly, the physical world provides a much larger and richer source of information for training than any dataset or even most simulated worlds can provide.
However, the data from the physical world are poorly aligned with the assumptions commonly made by most learning techniques. For example, learned classifiers often assume that the data are independent and identically distributed (IID) between training and test regimes. The assumption of IID data is critical to much of modern machine learning theory, but is fundamentally untrue of the physical world. Embodied agents live in a non-stationary, partially observable world where the data might be correlated through an intractably large number of latent factors. The data distributions are constantly changing, requiring the agent to learn over time what variations in the data might simply be the result of perceptual noise (or aliasing), versus changes resulting from its own actions or the interventions of other agents, versus variations that might represent a fundamental change to the environment.

Thirdly, embodied agents cannot assume that their goals, specifications and rewards are fixed for all time. Following \citet{Simon1956,Simon1996}, it is useful to consider an embodied intelligent agent in terms of the range of tasks the agent must accomplish, the environments it must operate in, and how the tasks and environments affect the architecture of the agent's intelligence. Another of our conjectures is that a good architecture for an embodied agent is one that generalizes across tasks and environments, which implies that the agent's representation of tasks and specifications must allow generalization and rapid adaptation to new tasks and environments. Reinforcement learning and decision theory have historically represented tasks and goals through rewards and loss functions. However, reward functions are not easily adapted to substantial changes in the environment, nor are they easily adapted to more complex changes to the task than simply varying the goal state, and in fact may be an inefficient representation for allowing generalization.

Fourthly, the fact that the environment and tasks can change has implications for how the agent learns. While embodied agents may have access to orders of magnitude more data, the distribution of data at any one time is very local, and drawn from a much smaller measure than the true distribution of data the agent can encounter over its lifetime. An embodied agent must recognize the need to act to acquire specific kinds of data, 
not only to perform well at the current task, but also to build a rich enough theory of the world to allow the agent to generalize to future tasks and future environments.  It may be precisely the feedback loop between action and inference that allows an embodied agent to learn a model that can make reliable predictions of the effects of actions, rather than merely learning to approximate the functional relationship between input and output.



Finally, the morphology of an agent is itself a decision variable for embodied intelligence \citep{lakoff1980metaphors,shapiro2007embodied}. What senses are available to the agent, the degrees of freedom of its actuators, the specific power available to the agent, all have tremendous implications about what the agent \textit{can} learn about the world and what actions the agent can decide to take.  The morphology of an agent also influences what the agent must learn about its environment, as the concrete embodiment can encode inductive biases that greatly facilitate learning.

\begin{figure}[t]
	\centering
	\includegraphics[width=.99\textwidth]{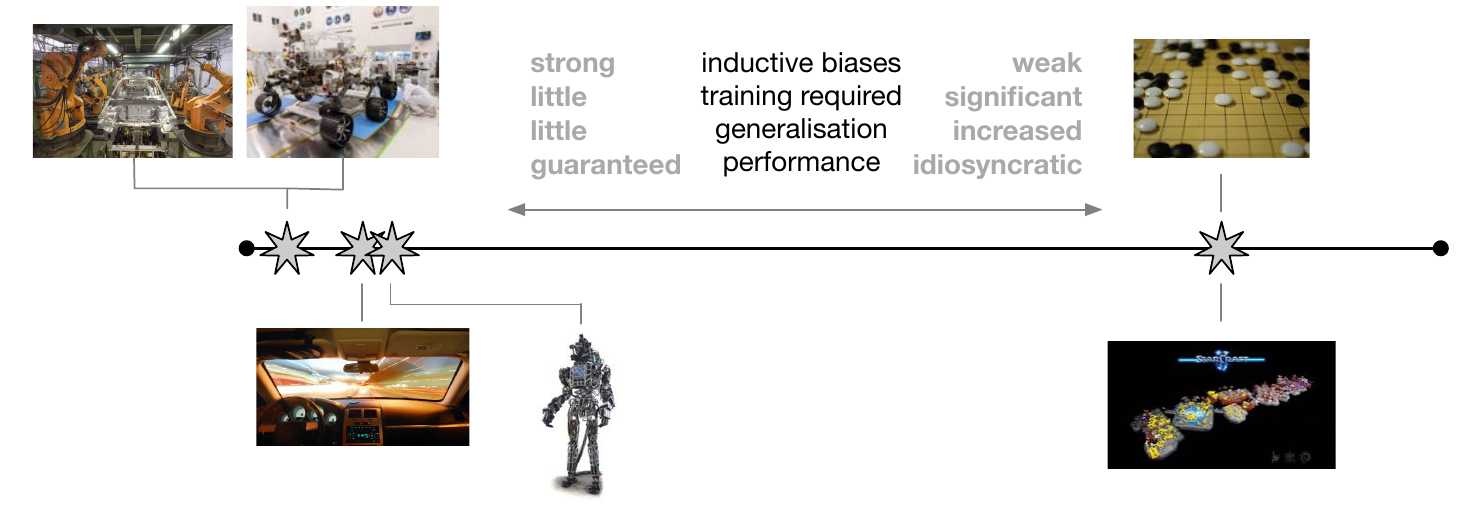}
	\vspace{-.2in}
	\caption{An \emph{agent-based} view of the spectrum of \emph{machine} learning. At one extreme (far left) everything is known. No learning is required, and performance is rigid but can be guaranteed via techniques from traditional systems engineering. At the other extreme (far right) nothing is assumed known and therefore everything must be learned from very large corpora of data. At this end, in theory, a system is able to generalize extremely well from one task to another. While approaches such as \emph{MuZero} \citep{schrittwieser2020mastering} have demonstrated the art of the possible with rather weak biases, to date such successes have mainly been confined to the context of game environments. In reality, AI agents commonly live in the continuum between these extrema. Current successes in real-world \emph{embodied} agents remain focused on specific tasks such as factory automation and autonomous driving while general learning techniques have not yet been shown to succeed on real world robotics tasks to a similar degree. We posit that current advances in AI technology do not lend themselves readily to advancing embodied agents towards the right-hand side of the spectrum. As research pushes towards the safe deployment of fully autonomous vehicles as well as the development of more versatile articulated robots, bridging this divide poses both challenges and opportunities for \emph{robot} learning.  
	}
	\label{fig:spectrum}
	\vspace{-.2in}
\end{figure}


The success of many robotic systems has been driven by the ability to constrain the operational environment and the set of tasks, avoiding many of the challenges described above. The constraints of a given environment and task enable an \emph{inductive bias} in the model and the learner, and we can consider a spectrum of inductive biases (see Fig.  \ref{fig:spectrum}). At one end of the spectrum, the environment and task are so constrained that a very strong inductive bias is possible, allowing learning to happen quickly and reliably from small amounts of data.  The price we pay for such success is a high degree of specialization in the resulting system, with little generalization.  Towards the other end of the spectrum, the inductive biases are significantly weaker (or stronger inductive biases remain elusive), leading to greater flexibility of the learner but also requiring increasing amounts of training data. At this end, in the limit, nothing is known and everything has to be learned. In reality, roboticists commonly find themselves somewhere between these extrema.  We may assume \emph{some} prior knowledge, but we can consider a system truly autonomous when it can function long enough such that almost every prior assumption is eventually violated. In poorly understood, data-impoverished, highly dynamic, potentially adversarial domains such as disaster response, planetary exploration or even everyday life, the need for a mature theory to enable embodied agents to learn correctly, efficiently and safely is critical.  A central challenge is therefore how to choose an architecture with an inductive bias of the learner that generalizes across a range of unknown and changing tasks and environments, and matches the requirements of physical constraints described above.

We begin with an examination of inductive biases for embodied intelligence and then move on to exploring some of them in more detail. In particular, we pose the following guiding questions for our discussion:
\begin{itemize}
    \item What are the appropriate inductive biases that allow an embodied intelligence to learn most effectively while being robust to changes in task and environment? We conjecture that the necessary inductive biases for embodied learning are based in action and perception, and we address this conjecture in section \ref{sec:inductive-bias}. 
    \item How do we design the  architecture of an embodied intelligence such that it can learn effectively and act robustly? We conjecture that an essential ingredient of such an architecture is the ability to effectively amortize reasoning while maintaining the introspection and capability to perform more elaborate inference on demand. We address this conjecture in section \ref{sec:fast-slow}.
    \item What is a representation appropriate for specifying both an embodied agent's model and the task to be performed? We conjecture that a specific form of compositionality is required of the representation that is absent from many architectures, and we address this conjecture in section \ref{sec:logic}.
    \item To what extent does the morphology of the agent affect its ability to learn? We conjecture that physical design is significantly complemented by effective learning methodology and should not be treated in isolation.  We discuss this conjecture in section \ref{sec:hardware}. 
\end{itemize}

Finally, an open question in embodied intelligence is how these systems should be evaluated. At heart, how do we as a community know if progress is being made? We conclude with a discussion of the challenges of not only evaluating but verifying an embodied intelligence that can learn. 

\section{Inductive Biases for Embodied Intelligence}
\label{sec:inductive-bias}
It is well understood that any learning system must have an inductive bias in order to make predictions outside the training set \citep{wolpert1997no,wolpert2002supervised}. An inductive bias is formally ``any basis for choosing one generalization over another, other than strict consistency with the observed training instances'' \citep{mitchell1980need}. Learning for embodied agents is no different in requiring inductive biases, however, we are not the first to observe that the nature of the  bias may be very different for an embodied agent \citep{thrun1995lifelong,burgard2020perspectives,kaelbling2020foundation}. While traditional inductive biases in machine learning, such as minimum description length or weight sparsity, are useful for embodied intelligence, additional architectural, algorithmic or even learned biases are increasingly moving into focus for embodied agents.

\subsection{Limitations and Challenges} 

It is useful to consider how the challenges of learning for embodied agents has implications for inductive biases. Our first challenge is to learn safely, given potential risks to the learning agent, which requires a theory of how to order models or hypotheses in a way that minimizes risk to the agent.  The field of human cognitive development provides one possible such theory in the ``zone of proximal development'' \citep{Vygotsky78}, which is often used to explain how children explore and learn. Vygotsky defined the zone of proximal development as ``the distance between the actual developmental level as determined by independent problem solving and the level of potential development as determined through problem-solving under adult guidance, or in collaboration with more capable peers'', essentially the set of tasks that are just beyond what a child can do, but the child can do with help.  There are a number of reasons why learning in the zone of proximal development is most effective for teaching children, but an important part of learning in the zone of proximal development is that it minimizes risk to the learner. The notion of scaffolding --- a process by which a teacher helps the learner within the zone of proximal development --- exploits this idea for effective learning.

To address our second challenge of learning when the world is non-stationary, the data are not identically distributed, and the agent encounters data that are very different from its training set, the agent must be able to generalize.  \citet{fodor1988connectionism} defined the concept of a ``systematic capacity'' as one where ``the ability to produce / understand some sentences is intrinsically connected to the ability to produce / understand certain others.''  This in itself is not a sufficient inductive bias but \citet{bahdanau2018systematic} broadened the idea to ``systematic generalization'', that is ``the ability to learn general rules on how to compose words''. The ability to learn general rules that lead to a causal theory of the world is important to an embodied agent because in a non-stationary, non-IID world, the training data will never be sufficient to represent all future queries. We conjecture that an inductive bias that encodes systematic generalization by encouraging a set of general rules to be learned may be the best way to enable agents that are robust to variation in task and environment. We will revisit this notion more explicitly in section \ref{sec:logic}.

Embodied agents must also be able to manage the computational growth that can result from learning in an open-ended environment. An effective approach is to reason at multiple levels of abstraction, making local decisions with highly precise models of the environment and the agent's dynamics when necessary, but equally being able to reason globally using much more abstracted representations \citep[e.g.,][]{toussaint2015logic,18-toussaint-RSS,LIS218,LIS278}. The use of hierarchical abstractions in decision making and task execution has been an active area of research for decades. Fueled by advances in deep learning, abstraction continues to draw increased attention, predominantly in the context of reinforcement and imitation learning \citep[e.g.,][]{Nachum-2018,le2018hierarchical,tirumala-2019}. However, hierarchical abstraction has not yet been used substantially in learning for embodied agents, with some modest exceptions  \citep[e.g.,][]{sutton1999between,dietterich2000hierarchical,levy2017learning,konidaris2018skills}. The state of the art in most operational embodied agents is to hand-code an abstraction, and even the most carefully hand-coded representations are brittle in the face of a non-stationary world. A useful inductive bias for an embodied agent may be an ability to learn its own abstractions which balance the computational efficiency of approximation induced by the abstraction against the potential for loss in performance. Recent works on state-space learning and control serve as examples  \citep[e.g.,][]{watter2015embed,karl2016deep}, as well as the increasing emergence of work on composable, object-centric deep generative models \citep[e.g.,][]{eslami2016attend,greff2016tagger,kosiorek2018sqair,burgess2019monet,greff2019multi,von2020towards,jiang2020scalor,engelcke2020genesis,nguyen2020blockgan}.  

A final requirement of embodied learning is to recognize that the agent has a finite amount of available energy or specific power. While some inductive biases such as minimum description length will tend to favor models that are more energy efficient, there is a need for a more substantial investigation of how an agent's energy limitations can form inductive biases. 

\subsection{Opportunities and Future Directions}
    
Given these challenges, there are a number of possible directions which may yield inductive biases that enable richer forms of learning for robots and other embodied agents. Three directions, in particular, would address the requirements and advance our understanding of choosing appropriate inductive biases for embodied agents that are learning.

A first possible form of inductive bias is to leverage the idea from cognitive science known as the ``core knowledge hypothesis'' \citep{spelke2007core}, that human intelligence rests on four systems that are hardwired for specific reasoning tasks, have innate representational capabilities, and innate limits. The four systems are designed to reason about objects, actions, number, and space\footnote{\citet{spelke2007core} speculate that a fifth system may exist for reasoning about ``potential social partners and social group members'', which would match the requirement that embodied agents recognise they exist in the presence of other dynamic agents with goals and intentions.}.
More generally, there are sets of concepts that seem universally useful to an embodied intelligence, such as physical properties (e.g., masses, inertias, rigid lengths etc.) and physical laws (e.g., Newton's laws, the interaction of light and sound with matter, the interaction of electromagnetic radiation more generally with matter, etc.) that should guide the learning process. Having a strong inductive bias that prefers models that are consistent with physics would seem to be an extremely useful property. Implementing an inductive bias with a realistic model of physics may however be computationally burdensome and likely incomplete, which raises questions of how accurate a model needs to be to be useful, and how strong the bias should be.  Behavioral studies of human physical reasoning suggest that people draw on a noisy, approximate ability to simulate interactions between physical objects, that qualitatively coheres with mechanics but departs from it in quantifiable ways \citep{sanborn2013reconciling, battaglia2013simulation}.
Prediction accuracy may also be a key bias in developing an intuitive understanding of the world and shaping the underlying model. \citet{ha2018world} explore this bias in the context of training an agent to act in a learned world-model and find that the agent is able to learn in a model that is \emph{good enough} --- but not perfect.


A second form of inductive bias may be in the form of structural biases that drive abstraction.  While some disciplines such as natural language processing have had success by reducing their explicit priors over internal representations and structure, there is not yet substantial evidence that embodied intelligence will succeed using an inductive bias that is purely ``signal-to-symbol'', where the internal representation is a quantization of the input \citep{bajcsy1995signal} --- a structural bias that is more than merely quantizing the input is needed. Reasoning about action is imbued with the study of dynamics, and the basins of attraction arising from the consequent energy landscape offer at least one source of effectively grounded symbols --- their systematic study launched the field of topology.  Systematic generalisation suggests that model composition must be a key element for embodied intelligence, and the prospects for a compositional language of basins appears bright \citep{Koditschek2021}.  

There is a lack of general consensus as to whether neural networks themselves are composable, at least in part because there is a lack of consensus on the definition of composition \citep{hupkes2019compositionality}. Nevertheless, with relatively few exceptions, current learning approaches for learning composition functions do not perform well when tasked with learning the kind of composition that abstract reasoning systems excel at --- for example, extrapolation outside the training set or inference dependent on global context. We will revisit the question of how to introduce compositionality as a structural inductive bias within a learning agent in Section \ref{sec:logic}. 
The compositionality question is related to inductive biases about causality.  Causal knowledge describes underlying mechanisms that can be composed to explain observations, such as the laws of physics, and is related to System 2 inductive biases~\citep{goyal2020inductive} discussed in the next two sections.

A third and implicit form of inductive bias may result from how data are curated and presented to the agent. For example, an ordering over tasks and environments, such as a curriculum, induces an ordering over the concepts to be learned, which often smooths the optimization path during learning but can also create an implicit bias in what is learned. Curricular learning has been shown to be an effective form of meta-learning \citep{finn2017model,narvekar2017autonomous}, and is very much related to the ideas from developmental cognitive psychology such as the zone of proximal development that enable efficient and safe learning. However, while there have been recent efforts in addressing this challenge, e.g., \citet{botvinick2019reinforcement}, there is not yet a principled theory of the relationship between meta-learning for embodied agents and the corresponding inductive bias, nor how to choose a curriculum that supports the different requirements of an inductive bias for embodied intelligence.

Finally, there are dangers of having inductive biases that are too strong. For many conventional machine learning applications, there are principles that can provide guidance in choosing an inductive bias appropriately. For an embodied intelligence that is attempting to learn from small amounts of data that are neither independent nor identically distributed, and when the size and scope of the domain cannot be known ahead of time, it is important to understand the risks and tradeoffs of different inductive biases. 


\section{Robots Thinking Fast and Slow}
\label{sec:fast-slow}

As we have already argued, when it comes to agents that can act and interact, many of the advances in AI have played to the strengths of virtual environments: infinite training data is available, risk-free exploration is possible, and acting is essentially free. In contrast, we require our robots to robustly operate in real-time, to learn from a limited amount of data, take mission- and sometimes safety-critical decisions and even display a knack for creative problem solving.  Cognitive science suggests that, while humans are faced with similar complexity, there are a number of mechanisms which allow us to safely act and interact in the real world. In addition to concepts such as the core knowledge hypothesis and the zone of proximal development that represent promising inductive biases, we focus in this section on a particular set of architectural  biases inspired directly by Dual Process Theory (DPT). Popularized by Daniel Kahneman's book \emph{Thinking Fast and Slow} \citep{Kahneman2011}, DPT postulates that human thought arises as a result of two interacting processes: System 1, an unconscious, intuitive response system, and System 2, much more deliberate reasoning.
If we accept that Dual Process Theory plays a central role in our own successful interactions with the world, we can explore a similar approach towards realising robust, versatile and safe embodied intelligent systems. 




A key observation of DPT is that faced with an every-day challenge like game-play, driving or stacking plates, we usually do not explicitly analyse the governing laws of the particular process. Instead we tend to simply act according to what the situation demands, as informed by our senses, based on intuition or even an innate reflex-like behaviour. The advent of deep learning has afforded our agents principally two things: (i) an ability to learn arbitrarily complex mappings from inputs to outputs; and (ii) an ability to execute these mappings in constant time. These, together with an ability to learn structured, task-relevant embeddings in an unsupervised manner, afford researchers a different view on the computational architectures they employ. The ability to learn complex mappings endows our agents with an ability to perform very complex tasks at useful execution speeds. Direct human supervision, reinforcement learning, task demonstrations, complex learned models as well as the increasingly popular concept of system-level self-supervision all fit into this narrative. In game play, DeepMind's AlphaGo Zero \citep{silver2017mastering} as well as the closely related \emph{Expert Iteration} algorithm \citep{anthony2017thinking} distill knowledge from Monte Carlo Tree Search (the oracle) and self-play into a model which predicts value and probability of next move given a particular board position. In robotics, OpenAI's Learning Dexterity project \citep{andrychowicz2020learning} distils knowledge gained in simulation through reinforcement learning and domain randomisation (the oracle) into a policy which can control a Shadow Hand to move an object into a target pose. In the context of autonomous driving, \citet{BarnesICRA2017} distill, via the automatic generation of training data, hundreds of person-hours of systems engineering into a neural network model which predicts where a human might drive given a particular situation. Recent work on intuitive physics learning distills data arrived at through physical simulation into neural network models, see, for example, \citet{wu2015galileo,lerer2016learning,wu2017learning,li2017visual,groth2018shapestacks,janner2019reasoning}.
Owing to their ability to mimic the expertise of an oracle in a time- (or generally resource-) efficient manner, one might view the execution of a neural network model as analogous to an efficient, \emph{intuitive} System 1 response. 

At the same time, roboticists and AI researchers have spent decades developing System 2 equivalents. Symbolic reasoning, traditional planning approaches and even simplistic but time-intensive brute force methods, all constitute deliberate and often effortful task solvers. Robust, real-world performance thus seems to require computationally efficient policies empirically tuned to a particular task and environment \emph{as well as} more computationally intensive approaches capable of systematic generalisation in that they are robust to variations in the task and environment. 
In the view offered here, we may be able to leverage machine learning to distill more resource-intensive, deliberate System 2 responses into learned models which mimic these experts in an efficient and effective way to form an intuitive System 1 response. 

\subsection{Limitations and Challenges}

The narrative of distilling knowledge into rapidly executable neural network models allows us to achieve significant, often game-changing, computational gains. However, as roboticists we are still faced with a substantive and foundational challenge when it comes to applying machine learning systems in the real world: the routine violation of the various assumptions made by our systems. As discussed in section \ref{sec:introduction}, an embodied intelligence cannot assume the data are independent and identically distributed. In contrast, as robots operate over ever longer time scales in increasingly unstructured environments, the data encountered significantly deviate from the training distribution. Additionally, we expect the embodied intelligent systems to generalize to unforeseen tasks and environments.  In practice, together with the approximate nature of our algorithms, these assumptions lead to learned models which are often over-confident and whose performance can only loosely be bounded (if at all) using traditional methods \citep[e.g.,][]{grimmett2016introspective,richter2017safe,NIPS2019_8981}. The result is that our robots lack the ability to reliably know when they do not know and take appropriate remedial action. Despite many attempts over the years at remedying this shortcoming \citep[e.g.,][]{settles2009active,hsu2010algorithms,li2011knows,pentina2014pac,NIPS2017_7154} we are still no closer to a practicable solution. Our conjecture is that a Dual Process Theory perspective may well provide a way forward to address this challenge.  



A second limitation of existing approaches is demonstrated by the considerable evidence that humans represent and use estimates of uncertainty for neural computation in perception, learning and cognition~\citep{deroy2016metacognition}. However, how \emph{metacognitive} uncertainties are derived and utilized is only gradually being discovered. Special, metacognitive circuitry in the human brain suggests knowledge integration above and beyond raw perceptual signals~\citep{deroy2016metacognition}. The \emph{Feeling of Knowing} process, for example, enables humans to effectively choose a cognitive strategy (e.g., recall vs. reasoning) likely to succeed in a given circumstance  \citep{reder1992determines}. Moreover, recent work on \emph{multi-sensory} perception suggests that metacognition is instrumental in discovering \emph{causal} structures in order to form a coherent percept from multi-modal inputs \citep{deroy2016metacognition}. 

Finally, while we may now have a technical blueprint for components on either side of the systemic divide, much uncertainty remains around the nature of the cognitive processes involved. Similarly, a strict categorization of the complex landscape of inter-operating neural processes into two types of systems as proposed in DPT is widely recognized to be a significant oversimplification. Nevertheless, it serves as a conceptual starting point. How to design an architecture that effectively combines the best of both worlds remains an open and potentially fruitful research question~\citep[e.g.][]{goyal2020inductive}. It is not clear whether components of both systems run in parallel or run on demand with an explicit handover between deliberate planning and low-level intuitive policies. In analogy with metacognitive challenges faced by humans, \emph{when} a System 1 response is appropriate over a more deliberate deployment of System 2 remains one of the open questions to be addressed. In humans, these two forms of processing interact in that System 2 can suppress, inform and even train System 1 responses \citep{Kahneman2011}.  Furthermore, it is not clear if the architecture is two separate systems or is in fact a continuum or tight integration of processes capable of fulfilling either part.


\subsection{Opportunities and Future Directions}
These limitations and challenges immediately point at a set of now viable technical approaches in which the outcome of a downstream system (either in terms of success/failure or in terms of confidence in outcome) given a particular input is distilled into a machine learning model\footnote{Statistical outlier detection also falls into this category.}. Predictive models of performance are now relatively common-place in the robotics literature. They have a long-standing track record in predicting task success in manipulation and complex planning tasks \citep[e.g.,][]{pastor2011skill,kappler2015leveraging,krug2016analytic,pinto2016supersizing,morrison2020learning} and are increasingly used, for example, to predict the performance of perception and vision-based navigation systems \cite[e.g.,][]{guruau2016fit,daftry2016introspective,dequaire2016off}.



Human intelligence is far more robust and uncertainty-aware than our best learning systems and operates in significantly non-stationary (in the statistical sense) environments. We draw on a rich capacity for \emph{metacognition} \citep{cox2005metacognition,deroy2016metacognition}: the process of making a decision, the ability to know whether we have enough information to make a decision and the ability to analyze the outcome of a decision once made. A predictive neural network can be injected with dropout noise~\citep{gal2016dropout} or another trained to predict the magnitude of out-of-distribution errors in these predictions, and thus estimate epistemic uncertainty~\citep{gurau2018dropout,jain2021deup}, which can be used to guide decision-making and exploratory behavior. One of the interesting aspects of a Dual Process Theory for robots is the fact that metacognition may find a natural place in such a construct: the open question is whether a theory of metacognition can be used to bridge the two systems.

Another set of viable technical approaches draws on new ways of combining causally structured generative models and Monte Carlo inference with machine learning. Causally structured models and Monte Carlo inference have a long history in AI and robotics, precisely because, taken together, they offer a way to model the causal structure of non-stationary systems, perform data-efficient parameter estimation, and build inference processes for state estimation that ``know when they do not know''. Unfortunately, it has not been easy to scale up inference in causally structured models such that it is possible to track large, complex environments in real-time. One approach is to use slow, offline inference to fit the generative models to data, yielding realistic simulations that can then be used to generate fully labeled, synthetic training data for bottom-up neural networks. Additionally, one can retain a lower-resolution structured generative model ``in the loop'' at inference time, and combine fast, bottom-up learned Monte Carlo updates from neural networks with slower top-down, model-based Monte Carlo. Recently, new probabilistic programming languages such as Gen \citep{cusumano2019gen} have been developed that make it much easier to write structured generative models and carry out real-time, approximate Bayesian inference via custom hybrids of neural, symbolic, and Monte Carlo methods. These approaches have been applied to solve 3D scene perception problems \citep[e.g.,][]{kulkarni2015picture}, and can run in real time while producing more robust results than approaches that only depend on learning. One benefit of ``model in the loop'' architectures is that they use the posterior probability density under the causal generative model to assign quantitative (relative) confidence levels to the output of machine learning algorithms. Similar architectures might be applied to navigation and planning tasks that build on the outputs of perception, using top-down model-based reasoning to quantify confidence in the outputs from fast, bottom-up learning.

A tantalising opportunity lies in System 1 computation  helping to solve one of the key computational limitations of symbolic AI systems, i.e., that of the intractable cost of search: deep generative models~\citep[e.g.][]{kingma2013auto,goodfellow2014generative,bengio2021flow} may play a role similar to imagination in that they can be trained to produce good candidates for search or planning consistent with more expensive System 2 sequential reasoning steps~\citep[e.g.][]{ha2018world}.

A cognitive science theory related to the DPT is the Global Workspace Theory (GWT) from~\citet{baars1993cognitive,baars1997theatre}, which postulates a communication bottleneck between specialized brain modules, with these expert modules competing for being able to send content through this bottleneck to be broadcast to all the modules. The connection with the DPT is that this bottleneck is the consciously accessible working memory which humans can report verbally, i.e., corresponding to System 2 content. However, the details of the computations performed inside the competing expert modules is not consciously accessible (and probably too complex to be completely verbalizable) and corresponds more to System 1 machinery. An interesting question  is the purpose of such a strong communication constraint. \citet{bengio2017consciousness} and \citet{goyal2020inductive} hypothesize that it induces an inductive bias that would make modeling the world at the abstract System 2 level good at capturing the kind of sparse dependencies and causal mechanisms that humans describe through language. The next section elaborates on the notion of multiple levels of abstraction and representation and the kind of discrete representations and sequential reasoning processes which humans tend to use at the higher System 2 level.


We close this section by noting that opportunities exist not only for addressing learning in embodied agents, but conversely also for advancing our knowledge in cognitive science. The latter often requires complex experimental procedures which, by design, need to disrupt the agent's learning process. Robotics, on the other hand, \emph{allows} the design and close inspection of the mechanisms involved in the learning process --- including control over individual components, the environment and modes of interaction. Much of this section makes the case that mechanisms already discovered in the cognitive sciences may cast existing robotics work in new light, with the aim of establishing a meaningful technological equivalent to the Dual Process Theory. In particular, they may provide a blueprint towards architecture components we are still missing in order to build more robust, versatile, interpretable and safe embodied agents. Conversely, as also noted by \citet{botvinick2019reinforcement}, the discovery of intelligence architectures which successfully deliver such dual process functionality may \emph{equally} provide fruitful research directions in the cognitive sciences.

\section{The Role of Logic in Embodied Intelligent Systems that Learn}
\label{sec:logic}

Having identified the need for systems for inference and decision making that can reason at different levels of abstraction, we turn our attention to the question of what representations enable learning at the different levels.  Learning representations for sensorimotor perception and control (i.e., System 1) is currently a well-studied problem as demonstrated by the vast body of work in deep representations for perception and control.  However, for deliberative inference and decision making, it is critical to consider different classes of representation languages.  A hallmark of problems that require deliberative reasoning is that they involve a form of dynamic compositionality over abstract concepts, long horizon and high dimensionality that may be more difficult to implement using System 1 machinery~\citep{lake2017building,bahdanau2018systematic}.  The strategy, in System 2, is to render these problems tractable by exploiting structure in the problem, including factoring into weakly interacting sub-components (possibly involving different aspects of the state space or different temporal sub-regions), and ``lifting,'' or abstracting over arbitrary sets of entities in the domain.  Classically, the languages that have effectively enabled compact representation and efficient reasoning in very large problems are {\em logics}, including propositional logic, first-order logic, AI planning domain description languages (PDDL), graphical models (which can be viewed as a probabilistic version of propositional logic), probabilistic planning languages, probabilistic programs, and temporal logics. Similarly, the tools of topology (and the computational efficacy of its algebras) hold a relationship to robotics analogous to that of logic relative to computer science. 

Models represented in any of these languages are much closer to a specification provided by a human engineer compared to reflex-based models represented as neural networks. However, the inference and learning questions for all of these modeling paradigms are not inherently different: in every case, the models can be used to make predictions or to determine action choices, and they can be fit to data by machine learning methods that endeavor to make the models' outputs match those in a data set.  Nonetheless, the algorithmic techniques for model fitting may be substantially different and the choice of what kinds of problems are best matched to what kinds of models is not well-understood. The choice of representation language therefore depends on the types of problems it is suited for and the underlying inductive bias --- does it fit the problem at hand, and thus substantially decrease the sample complexity for learning?  One important feature of many logical representations is {\em compositionality}, where parts of the model have independent semantics, allowing them to be learned independently and then composed in combinatorial ways to solve many different problems, providing a particularly aggressive and useful form of generalization. It is through this compositionality that we may hope to achieve the systematic generalization introduced in section \ref{sec:inductive-bias}.

\subsection{Limitations and Challenges}

Learning logical representations has been an active research area for a considerable time. One of the main focal points of learning in logical representations is inductive logic programming (ILP), with the goal of inferring a logical theory of the world that entails a given dataset. While the use of ILP has been examined previously for embodied intelligence \citep{bratko2010comparison} including mobile navigation \citep{leban2008experiment} and learned grasping \citep{antanas2015relational}, these results have not demonstrated the same level of success as alternate approaches. How to learn compositional rules that represent abstract actions such as learned STRIPS or PDDL representations of actions has been a subject of investigation from the early days of Shakey \citep{fikes1972learning} to more recent results \citep{konidaris2014constructing,sammut2015robot}.  However, much of the prior work assumes deterministic and often propositional representations. The ability to learn stochastic, factored and lifted models of both perception \citep{icra14ensmln} and action models \citep{mugan2011autonomous,kruger2011object} is crucial for operating in a physical world. In particular, \citet{pasula2007learning} argue that propositional, relational and deictic learning represent different inductive biases, and ``deictic learning provides a strong bias that can improve generalization''.  Unfortunately, learning logical representations through ILP often suffers from an exponential growth in complexity with the size of the theory to be learned, and also can have difficulty with noisy or inconsistent data.  Recent progress has shown the effect of neurally-inspired induction, such as differentiable ILP \citep{evans2018learning,Payani2020}, that can deal effectively with noisy data.  Neural logic machines \citep{dong2018neural} have shown how to recover a set of lifted rules in very simplified physical environments while avoiding problems of scalability.  Nevertheless, the application of any kind of ILP or learning logic machines to embodied intelligence has been limited in scope and represents a substantial open challenge.

Another limitation of most existing work in learning logical representations is the reliance on an initial theory of the world, often referred to as ``background knowledge'', comprising an initial set of predicates and propositions with truth values that are either known \emph{a priori} or can be inferred from observation.  In very simplified domains such as family tree reasoning (a common benchmark in the ILP community, \citealp{wang2015soft}) or blocks world, it may be reasonable to write down an initial background knowledge theory that is sufficient to support induction of a complete set of predicates and rules.  However, the background knowledge is ``similar to features used in most forms of ML'' \citep{cropper2021inductive}.  Just as deep learning has substantially reduced the need to handcode feature functions for many problems, it is possible that the current reliance on handcoded background knowledge theories is a substantial limitation in learning logical representations for embodied intelligent systems. Techniques that get closest to learning symbolic representations from completely continuous, sensorimotor representation still rely on background knowledge of motor primitives or skills \citep{konidaris2018skills}. At the same time, as discussed in section \ref{sec:inductive-bias}, the ``core knowledge hypothesis'' described in section \ref{sec:inductive-bias} would seem to suggest that biological systems themselves depend on background knowledge. It is an open question whether background knowledge is a limiting inductive bias for embodied intelligence, or whether more research is needed to identify the \emph{correct} background knowledge for an embodied intelligence.

Our notion of a learning embodied agent is one that will eventually need to represent concepts that could not have been anticipated ahead of time. Being able to grow the primitive logical representation from the raw sensor data, a process sometimes referred to as ``symbol emergence'' \citep{taniguchi2018symbol} is an important ability for an autonomous agent. There has been work on attempting to learn a discrete planning domain model from completely continuous, sensorimotor representations \citep{Asai2018,Asai2019,Ames2018}. Additionally, some previous work has demonstrated that these approaches in learning discrete representations for actions can be coupled with a convolutional neural network used for processing visual input, to create an end-to-end system that learns both a symbolic representation of the visual input and a logical theory of actions for planning in toy domains such as Sudoku \citep[e.g.,][]{wang2019satnet,dong2018neural}. Similarly, previous work has demonstrated that a partial representation can be learned ahead of time, and then recruited by a reactive system that deforms real-time sensory instances into their learned topological model via real-time change of coordinates \citep{vasilopoulos2018sensor_a}. However, it is not well-understood how to scale up these initial results to more complex domains encountered by an embodied intelligent system.  

Furthermore, in order to obtain the advantages of logical representations such as composition, the inductive biases of the learning process must not only satisfy the physical constraints of an embodied intelligence, but also meet the needs of the logical inference process. For example, in factoring, a decomposition must be found that renders different aspects of the problem relatively independent.  In lifting, objects must be ``reified'' in a way that allows useful abstractions.  This search for the fundamental discrete structures requires an inductive bias towards factoring the world into independent and composable pieces, for example as might be formalized by a type theory of energetically grounded symbols \citep{Koditschek2021}. 

Lastly, even with a well-defined discrete representation, an embodied intelligence exists in a continuous environment, and there must be a connection between the abstract, usually discrete, representation to the concrete, usually continuous input and output signals. This connection between the abstract representation and the physical world is the ``symbol grounding problem'' \citep{harnad90}. However, existing models of symbol grounding fail to meet nearly all the challenges faced by embodied agents \citep{coradeschi2013short}; for example, learned models of symbol grounding again assume the data are IID, and that there is a fixed and known alphabet of symbols to be grounded to the input and output signals. The inductive biases in the current approaches to learning symbol grounding do not yet have principled techniques for determining when the world has changed and the corresponding learned model of symbol grounding has changed. In recent years, approaches such as unsupervised learning for option or skill discovery~\citep{gregor2016variational,eysenbach2018diversity,bagaria2019option} have attempted to learn primitives and discrete-level skills to incorporate compositionality for tasks such as long-horizon planning. While still preliminary, these works  put forward a promising direction for instituting some of the benefits of logic into neural models.

\subsection{Opportunities and Future Directions}

Developing new techniques for inferring the base or primitive representation that allows for compositionality is one of the primary open questions for a learning embodied agent. There is increasing evidence that the internal layers of many neural networks can encode compositional symbols in a way that relates to discrete logical representations. For example, ``disentanglement'' techniques can be used to force a learned representation that is factored, and results so far reveal surprisingly intuitive structure about the learned representation \citep{higgins2018towards}. These results suggest a path forward to learning the underlying abstraction that is the foundation of logical reasoning, although this work has not yet led to substantially improved performance in inferring compositional representations.

Opportunities exist for new techniques in learning lifted models, and graph neural networks (GNNs) provide a possible mechanism for neurally-inspired lifted models, in the sense that a model of fixed dimensionality can be learned and then applied to domains with arbitrary numbers of objects.  Neural networks embedded within graphical models \citep{krishna2017visual,armeni20193d} or full GNNs have been applied to scene perception and entity abstraction \citep{veerapaneni2020entity,qu2019gmnn}.  However, GNNs have not been shown to capture the power of quantification, and with the exception of some promising preliminary work \citep{simonovsky2018graphvae,franceschi2019learning,alet2019neural}, the assumption is that the graph structure is known ahead of time.

There does already exist an expressive and reasonably complete language that is composable and supports abstraction, in human natural language. There has been considerable work in symbol grounding efforts for embodied agents that relies on natural language to provide the set of symbols that are to be grounded \citep{tellex2011approaching,matuszek2013learning,thomason2015learning,paul2016efficient,patki2019inferring}, however, these approaches have not yet demonstrated the ability for a robot to acquire a large knowledge base without considerable human intervention and effort. The difficulty in much of this work is the need for highly-annotated, aligned corpora of data; there is an opportunity for new techniques that allow an embodied intelligence to acquire new symbols from language in a self-supervised manner. Attempts to learn symbolic or logical knowledge bases by, for example, reading the internet \citep{matuszek2005searching,NELL-aaai15,Alarcos19} have made more progress but remain incomplete and have not had substantial impact on embodied intelligence.

There is an open question as to whether or not background knowledge should act as a prior on the learner, as in the inductive biases discussed in section \ref{sec:inductive-bias}, or if in fact the learner should be attempting to derive its theory of the world from scratch using principles such as systematic generalization.  There is increasing evidence that a very plausible approach is to fix a simplified model ahead of time, e.g., a physics-based model, and then learn to correct for errors induced by abstraction \citep{ajay2018iros,zeng2020tossingbot}.  Alternately, there is evidence that embedding an entire physical model in a neural network and training it end-to-end may lead to models that are more robust and appropriate for the problem distributions they are faced with \citep{whiteson2018treeqn,amos2018,LIS249}. More recent techniques heavily rely on pretraining a universal model on large offline datasets---following a similar paradigm as universal language models such as GPT-3---which although is often not compositional can act as the background knowledge~\citep{levine2020offline}.

An interesting direction is to incorporate System 2 inductive biases in neural networks~\citep[e.g.][]{goyal2020inductive}. One starting point inspired by the GWT (see~\ref{sec:fast-slow}) is to construct a modular architecture where modules compete to be activated and communicate~\citep{goyal2019recurrent}. Using attention mechanisms to operate on sets of elements (rather than vectors) makes it possible to implement a working memory as the communication bottleneck~\citep{goyal2021coordination}, as in the GWT, yielding better out-of-distribution generalization. Forcing the messages exchanged between modules to be discretized using a shared vocabulary of abstract concepts yields further improvements to modular architectures~\citep{liu2021discrete}, including to the very popular Transformers~\citep{vaswani2017attention}. This kind of modular architecture can also be forced to process information through the sequential application of simpler rule-like but learned operations implemented by small expert MLP modules, yielding ``neural production systems"~\citep{goyal2021neural}. While these are inspired by classical AI production systems, all the rules are \emph{learnable} and can operate on distributed representations: end-to-end learning of an attention machinery can again be used to dynamically control, which rules are applied when, in what order, in order to minimize some training loss.

Finally, logical representations are often used in domains that require guarantees of correctness of the inference, which is especially useful in ensuring physical safety in many engineered systems and seem to have promise for ensuring the safety of an embodied agent~\citep{kress2009temporal,kloetzer2008fully,raman2015reactive}.  However, by construction, the learning process itself cannot provide anything other than statistical guarantees. As we shall discuss in section \ref{sec:real-robots}, for an embodied intelligence, statistical guarantees provide statements of robustness but not statements of safety; some external structure around the learner is required to ensure safety properties. There is an open question as to what kinds of learning can be applied to a formal representational language that preserves the ability to provide guarantees of correctness.

\section{The Impact of Morphology on Embodied Intelligence}
\label{sec:hardware}

A particularly strong but often unacknowledged inductive bias is introduced by the morphology of the robot. What sensors the agent possesses, what degrees of freedom it can act with, the dynamics of its motion, its own rigidity in interacting with the environment, the extent to which it \emph{can} interact with the environment all have a tremendous influence on what the agent can and cannot learn, and what computations can be ``offloaded'' to the body~\citep{hogan_impedance_1985,pfeifer2006body,muller2017morphological}.
However, physical agents are currently constructed from a fairly limited range of possibilities. The vast majority of vehicles rely on high-precision range sensing (typically based on lidar) and high-precision, stiff actuation. These design choices are motivated by the requirements of existing estimation, perception and control algorithms to have highly accurate models of the agent's sensors and actuators, and to have highly accurate knowledge of the agent's state and that of the environment around it. 

\subsection{Limitations and Challenges}

While the space of robots and embodied agents that have precise sensing and actuation include many useful types of platforms such as manufacturing robots, self-driving cars, unmanned air vehicles, and others, there are some limiting consequences to how robots are most often designed today. 
Robots with precise, stiff actuation tend to be either extremely slow, or have high energy, making it difficult and potentially unsafe for people to work alongside them. High-precision actuation is also typically expensive in terms of size, weight, power and cost and such robots are difficult to operate, limiting how widely they can be adopted by people interested in the questions of learning for an embodied intelligence. Highly precise robots constructed from rigid, articulated links generally must avoid contact with the environment except in specific and carefully controlled ways, which dramatically limits their ability to learn by interacting with the environment. This is one example of how the morphology of the robot impacts what kind of concepts can be learned. 
An interesting counter-example is represented by the low-cost manufacturing robots Baxter and Sawyer produced by Rethink Robotics. By virtue of low-energy series-elastic actuators, these robots were safe enough for people to work alongside, which drove their adoption by the robot learning community. These robots were also relatively imprecise, but it has been shown \citep{li2014localization,cremer2016performance,guan2018efficient} that an embodied intelligence can learn effective and accurate manipulation strategies despite the imprecision of the end effectors. 

A similar situation exists for sensing, in that the vast majority of operational autonomous robots rely heavily on laser range finders, which perceive the world in ways that are radically different from most biological systems. While laser range finders are extremely reliable in detecting obstacles and building detailed and dense geometric maps of the environment in order to plan collision-free motion, these sensors provide limited information about the visual appearance of the world (even when intensity information is utilized). There is a tremendous amount of appearance information embedded in the world  (signs, landmarks, etc.) that is inaccessible to a vehicle using only a laser range finder for navigation. While most robots now combine ranging with passive vision, each sensing modality is used to perceive very specific forms of information that are essentially orthogonal to each other. There is very limited understanding of how to trade off the different sensors, especially in a learning context. As with high-precision actuation, high-precision sensing is also expensive in terms of size, weight, power and cost. A successful counter-example is the Skydio drone, which demonstrates a surprising degree of navigation autonomy using a large number of cameras arranged in stereo pairs. However, the internal representation used by the Skydio vehicle is very similar to the dense geometric maps used by laser range finders, which creates an inductive bias in  not leveraging all semantic information at every representational level. 
Only recently have results begun  to emerge that use so-called ``semantic information'' for navigation \citep{civera2011towards,atanasov2014semantic,kostavelis2016robot} supported without a dense geometric model. The inductive bias induced by assuming that perception is provided by a range measurement system is another example of how the morphology of the robot impacts what kind of concepts can be learned. 

However, despite the evidence that the morphology and sensory modalities of an agent induce a substantial inductive bias in learning, there is relatively little understanding of exactly how morphology and sensor modality impact the ability to learn. For example, the tradeoffs between power, mass, sensing and computing are poorly understood, and especially how they impact learning. There is no principled way to modify the morphology to address a specific inductive bias. Similarly, we have limited understanding of how to design an embodied agent that can learn robustly across a specific range of tasks and environments; too often our agents are designed for one environment and immediately fail when presented with a slightly different environment or task\footnote{Many DARPA robotics programs have attempted to overcome this problem by requiring evaluation on a range of environments, such as the Fast Lightweight Autonomy program, which standardized the flight vehicle hardware but allowed sensor variability. The range of environments did appear to lead to performers choosing different sensor modalities that had different operational characteristics. Unfortunately, no sensor morphology was robust to all environments, signaling a technical gap.}.
    
\subsection{Opportunities and Future Directions}

Given these challenges, there are a number of possible directions where learning can be used to leverage a wider range of agent morphologies. Learning may be a way to reduce the costs of constructing and operating real robots. More compliant actuators and compliant bodies are often cheaper to build and safer to operate around people, and precision is only sometimes needed. Relatedly, robots can be designed to be more robust to situations likely to be encountered during online learning \citep{bhatt2021ihm}. Opportunities exist to design robots that learn to be precise only when necessary, using very different control strategies than are currently used, leveraging kinematic and dynamic constraints such as contact or inertia. Modalities such as sound provide a potentially rich and low-cost sensory stream that remains largely unexploited \citep{zoeller2020acousticsensing}. Constructing components from active materials represents an area where learning may be very useful in developing highly-capable control systems. Similarly, learning may enable richer forms of sensor fusion. An excellent example of how expanding our sense of plausible robot sensors through sensor fusion is the GelSight sensor \citep{li2014localization}, which uses visible deformations in gel at robot fingertips as a form of tactile sensing. The ability to interpret the deformations as force and contact information is enabled by modern machine learning techniques. Learning may also be the key to unlocking high-density, large-area tactile sensing using systems such as \citet{maiolino2013flexible}, which are increasingly emerging. Similarly, varying robot morphology and design may allow us to better understand existing limits in terms of energy consumption, computational complexity, reachability and observability.  

As discussed in the introduction, a substantial difference between conventional machine learning and learning for embodied agents is the availability of training data that matches the assumptions of the learner. Exploring the space of different morphologies, sensing and actuation paradigms is impractical when each new design must actually be physically fabricated and then used to collect data for learning. Simulators will be needed that are sufficiently high fidelity to enable learning to occur for robots that vary significantly from the current paradigms. Ideally, new classes of simulators could enable co-design of both the learning approach and the hardware itself. It has long been a central tenet of robot learning in particular that simulation is ``doomed to succeed'' \citep{Brooks1993}. However, the state of simulation, and consequently the value of simulation, is changing rapidly and the apparent limits of simulation may not in fact be fundamental to all simulations. Photo-realistic game engines are increasingly common-place and can enable embodied agents to learn from images. Physical simulators are also increasingly capable, although some physical phenomena are challenging to simulate in real-time such as deformations, friction effects and fluid effects. Nevertheless, with the advent of better, more physically realistic and photo-realistic simulation, new opportunities exist for learning in the context of new robot types. A combination of advances in simulators, safe learning, and sim-to-real methodologies is likely to enable designs with much wider variations in morphology, and, commensurately, insights into how morphology helps enable intelligent embodied behaviors.

\section{Assessing Robot Learning}

\label{sec:real-robots}




Embodied agents can have a very direct and physically damaging effect on their environment as compared to other applications that rely on learning-based methods, creating concerns of safety.  Verification, validation and benchmark-based performance evaluation are the cornerstones of safe deployment of active control systems and increasingly a core component of deployment of autonomous systems.  However, how to validate and verify the performance of an embodied agent that learns and adapts to novel experiences is an open question and it is likely that new principles for evaluating our embodied agents will be required.

\subsection{Verification and Validation of a Learning Embodied Agent} 

Conventional verification and validation techniques assume a predefined specification of the desired system behavior in an environment. \emph{Validating} the safety of a system means that no failure has been found after testing it under environment disturbances or when the probability of failure is below some threshold. Safety \emph{verification} means that a system is provably safe to all disturbances. Both validation and verification therefore require a model of the disturbances. For most realistic systems in robotics, such a specification and realistic disturbance model is nearly impossible to specify or sample from, for all but a very narrow set of tasks and environments. It may well be that the traditional requirements of verification and validation are in direct conflict with many of the goals of learning. Furthermore, the kinds of guarantees provided by most learning approaches are at best statistical, describing aggregate or asymptotic behavior rather than providing guarantees on any instantaneous query of a learned model. These kinds of guarantees also often rest on the previously discussed assumption that the training and test distributions are identical, which is hardly true in a real scenario.  Probabilistic guarantees may be extended when it is possible to determine how far a training set is from the current domain, task and environment (see section~\ref{sec:fast-slow}). Given these challenges, the successful verification and validation of a learning embodied agent will require novel approaches, based on innovative ways to characterize desired system behavior.


An interesting technical direction is the idea of adversarial testing. While adversarial examples have been used to demonstrate potential failures of learned models, it may be possible to automatically generate adversarial examples that test the response of an autonomous system. This is a very active area of research when it comes to adversarial images or other perceptual input  \citep[e.g.,][]{adversarial:2014,ranjan2019attacking,tu2020physically} but raises the question of how adversarial disturbances of an environment can be automatically generated that change in response to closed-loop behaviour~\citep{SinhaOTD20}. The nature of a proof of correctness of a learning system may be that no adversarial examples exist, at least in the context of a particular task environment. The challenge is not only to find such examples given the high sample complexity in realistic problems, but also to ensure the learning system is robust to these. \citet{corso2020survey} provide a broad survey of safety validation of black-box systems through simulation. These systems search for disturbances that cause the evaluated system to behave improperly and may use optimization, planning or reinforcement learning techniques for this search.


\subsection{Performance Evaluation of a Learning Embodied Agent} 

Evaluating the performance of, or ``benchmarking'', learning-based methods in robotics is similar in many ways to verification and validation. In benchmarking we are interested in how well such a system performs especially in comparison to other systems tested in similar tasks and environments. Benchmarking differs from the binary assessment of verification in that benchmarks can be used to measure progress during development, not just whether the final system meets a target level of performance.

Current benchmarking on {\em real robotics systems or data\/} can be approximately grouped into three kinds. First, there are anecdotal demonstrations of solutions on specific scenarios. These are often proofs of concept. Second, there are (offline) datasets, e.g., for pick and place~\citep{ArmFarm,mahler2017binpicking,pinto2016supersizing} or autonomous driving~\citep{Geiger2013IJRR,nuscenes2019,sun2019scalability} that often test specific perceptual skills such as inferring a robust grasp from visual data, accurate object detection or semantic segmentation in a street scene. Third, there are robotics challenges~\citep{AmazonPickingChallenge,DRC:2015:Atkeson,Righetti2014ARM,Eppner2018}, most prominently the DARPA challenges that often reveal how far we have come in a specific subfield of robotics: manipulation, navigation, driving, tasks in disaster relief scenarios, etc. 

Robotics poses a particular challenge for benchmarking learning-based solutions. A robot is a closed-loop system that acts and thereby influences the state of the environment, which in turn influences its next steps. This feedback loop usually prevents  the use of offline datasets for evaluation, as is commonly done for benchmarking of algorithms in computer vision or natural language processing. For example, the success of datasets like KITTI \citep{Geiger2013IJRR} and the more recent Argoverse dataset \citep{chang2019argoverse} have helped drive progress in estimation and perception for self-driving cars, but they have not yet had a similar impact on decision making.  
Major companies focusing on autonomous vehicles -- such as Waymo, Tesla and others, have publicly discussed the  importance of simulations and the significant amount of simulated driving that they are currently using for testing and technology development. Fully interactive simulations allow for both performance evaluation and learning. Furthermore, rare corner cases can be slightly modified to multiply interaction scenarios, allowing systems to be better tested and trained. In this context, the Open Source CARLA simulation environment \citep{dosovitskiy2017carla} and the challenges run by the CARLA team serve an important role for the broader academic and industrial research community.  

Simulation is a mechanism for systematic evaluation that does not require expensive and tedious deployment of robots into the real world. However, simulations for robotics have historically been notoriously low quality with a significant domain gap between simulation and the real world. Techniques such as so-called ``sim2real'' that use learned models to close the gap between the distribution of simulated and real data have shown some recent success in allowing embodied intelligence to learn effective control policies \citep{sadeghi2017sim2real,ramos2019bayessim,sim2RealDeepRL,sim2real_ws}; nevertheless, simulating realistic sensory data \citep[e.g.,][]{porav2018adversarial,yang2020surfelgan,weston2020there} is still a challenge and simulating contact is even harder. While many learning-based methods are developed and evaluated in simulation, it is unclear whether they would work in the real world. Examples of such benchmarks include Meta-world~\citep{yu2020meta}, Point Navigation in Habitat-AI~\citep{habitat19arxiv}  iGibson~\citep{shen2020igibson}, and the OpenAI Gym~\citep{BrockmanCPSSTZ16}. 
\citet{andrychowicz2020learning} presented a rare example of transferring a policy learned in this simulation environment to a real robot, but the success rate dropped by 50\% despite very large amounts of simulated training iterations on a randomized domain. A possible way forward is to consider ``real2sim" which attempts to identify a simulator that more closely resembles a few real world examples~\citep{SimGAN:2021}.

There has always been a conflict between functionality and generalization in robotics. Learning-based systems promise to generalize to unseen situations in many robotics applications, but when compared to methods that are application-specific (even with learning-based components), full learning-based methods often exhibit orders of magnitude worse performance on a specific benchmark. Therefore the question arises of what exactly are the correct metrics for a learning embodied agent --- it is an open question if the best metrics are performance on any given task, or if the metrics should characterize generalization over many tasks or to unexpected situations.
Specifically, Goodhart's law is often at play,  summarized by Strathern as ``When a measure becomes a target, it ceases to be a good measure'' \citep{Strathern97}. This phenomenon can be  seen in robotics challenges where the point is to win rather than to find solutions to problems that may generalize to larger variations of the given task. While some general insights have been gained through challenges (e.g., compliance for manipulation, suction cups for pick and place) these challenges often expose point solutions that address the specific goal of the contest but do not solve the actual problem of interest~\citep{DRC:2015:Atkeson,AmazonPickingChallenge}. 

It is imperative to develop new theories and metrics for evaluating embodied intelligence that lead to general solutions, rather than point solutions that are subject to Goodhart's law. Extending the ideas presented with respect to verification and validation~\citep{corso2020survey} such as adversarial testing for benchmarking is one promising direction. Another open question is how to systematically change evaluation metrics in a principled way to avoid overfitting and point solutions. While an ordered system of evaluations has typically not been considered by the robotics community, the area of meta-learning allows progress in this direction by re-using learned knowledge about the learning process itself to learn new tasks faster. 
Finally, it is important to note that  all of these approaches are linked by the notion of online learning, which raises concerns about the stability in terms of representation and models.

\section{Conclusion}

Embodied agents that are able to learn new representations, new models of the world and new skills will enable a new generation of robots and autonomous machines that can carry out a vast range of tasks over long time- and length-scales. At some point, the world no longer presents data from a stationary, independent and identically distributed test set. The agent must recognize the potentially unsafe consequences both to itself and to the environment around it as it acquires new data, builds a new model and tests the decisions it takes as a result of that model. The agent must also act to acquire that data, in the context of its own size, weight, and most importantly, energy constraints. 

Current systems that rely on the state-of-the-art in machine learning and artificial intelligence are reaching the limits of performance. We posit that these limits are imposed by the challenges described in this article. Remarkably, many of the same challenges of learning for embodied intelligence we describe here have not changed in 30 years \citep{Brooks1993}. Overcoming these challenges will require another step-change in both our understanding of, and the capabilities of, embodied agents and robots. 
In this article, we have argued that conventional machine learning and artificial intelligence do not adequately address the needs of an embodied agent that learns. There are technical challenges that must be addressed by the different communities building embodied intelligence systems. Our first hypothesis is that the inductive biases must be made explicit for embodied intelligent systems that learn, and in many cases new inductive biases must be developed to address the limitations of the current approaches to learning in an embodied intelligence. Second, we hypothesize that an embodied intelligence that learns must do so at multiple levels of abstraction, and claim that cognitive science theories like the dual process theory and the global workspace theory may in fact provide a path forward to a mature set of inductive biases enjoyed by humans and that may allow learning at such multiple scales and levels of abstraction. The absence of an explicit hierarchical representation of time and length scale is one limitation that has prevented robots from learning to perform complex tasks in stochastic environments. Third, we hypothesize that new languages are required for representing embodied world models at multiple time and length scales. In fact, we claim that some form of symbolic logic grounded in physical perception and action is essential for scalable representations that enable robots to learn and generalize effectively in the real world. Fourth, we hypothesize that robotics in particular has been too committed to specific morphologies, not only in physical form and actuation but also in sensing. While these actuation and sensing modalities have very much enabled robotic technologies in specific domains such as manufacturing or self-driving vehicles, these same modalities have limited our understanding of what embodied intelligence can really accomplish. Investing in new sensing and actuation modalities may create opportunities for robots with a very different set of capabilities, but such robots may depend heavily on learning for robust perception and control in ways that are not yet well-understood. Finally, embodied intelligent agents that learn  must do so safely, and we need ways to verify that safety. We need meaningful ways to compare different agents both in learning and in performance. Existing technologies, metrics and evaluation techniques are not adequate for a variety of reasons, including the risk of over-committing to specific standards and benchmarks. The entire point of learning is to be able to generalize to the unseen, even out-of-distribution, and it is currently very difficult to assess learning performance on novel tasks. 

Assessing these hypotheses and progressing our understanding of embodied intelligence is not a trivial task. However, our hope is that these hypotheses are sufficiently concrete to inspire new research directions and new investigations of embodied intelligence that are not currently underway. We look forward to future generations of embodied intelligent agents that are robust to a changing world, are robust to the assumptions of their human designers, and can operate in a populated world with true autonomy.

\begin{acks}
This paper is the results of a three day workshop sponsored by and held at Element AI, and their support is very gratefully acknowledged. Leslie Kaelbling assisted with several ideas in this paper, and her considerable time and help is also very gratefully acknowledged. 
\end{acks}

\bibliography{robot-learning}

\end{document}